\documentclass{article}

\pdfoutput=1

\usepackage{microtype}
\usepackage{graphicx}
\usepackage{subfigure}
\usepackage{booktabs} 

\usepackage{extarrows}
\usepackage{amssymb}
\usepackage{amsmath}
\usepackage{subfiles}
\usepackage{tikz}
\usepackage{pgfplots}
\usepackage{epstopdf}
\usepackage{soul}
\usepackage{enumerate}
\usepackage{mathtools}
\usepackage{todonotes}
\usepackage{sidecap}
\RequirePackage{algorithm}
\RequirePackage{algorithmic}

\newcommand{\interior}[1]{{\kern0pt#1}^{\mathrm{o}}}
\DeclareMathOperator{\relu}{ReLU}
\DeclareMathOperator{\rank}{rank}
\DeclareMathOperator{\Conv}{Conv}

\DeclareMathOperator{\sN}{\mathcal{N}}

\newtheorem{theorem}{Theorem}
\newtheorem{lemma}[theorem]{Lemma}

\newtheorem{corollary}[theorem]{Corollary}
\newtheorem{definition}[theorem]{Definition}
\newtheorem{remark}[theorem]{Remark}
\newenvironment{proof}{{\textbf{Proof\ }}}{~\hfill $\square$}

\usepackage[
	preprint,
]{nips_2018}

\usepackage[utf8]{inputenc} 
\usepackage[T1]{fontenc}    
\usepackage{hyperref}       
\usepackage{url}            
\usepackage{booktabs}       
\usepackage{amsfonts}       
\usepackage{nicefrac}       
\usepackage{microtype}      

\title{Analysis of Invariance and Robustness via Invertibility of ReLU-Networks}

\author{
  Jens Behrmann*\And S\"oren Dittmer*\And Pascal Fernsel\And Peter Maa\ss\AND
  Center for Industrial Mathematics\\
  University of Bremen, Germany\\
  \texttt{\string{jensb,sdittmer,pfernsel,pmaass\string}@uni-bremen.de} \\
}

\begin{document}

\maketitle

\begin{abstract}
Studying the invertibility of deep neural networks (DNNs) provides a principled approach to better understand the behavior of these powerful models. Despite being a promising diagnostic tool, a consistent theory on their invertibility is still lacking. We derive a theoretically motivated approach to explore the preimages of $\relu$-layers and mechanisms affecting the stability of the inverse. Using the developed theory, we numerically show how this approach uncovers characteristic properties of the network.

\end{abstract}

\section{Introduction}
\label{sec:intro}
\begin{figure}
\includegraphics[width=0.485\textwidth]{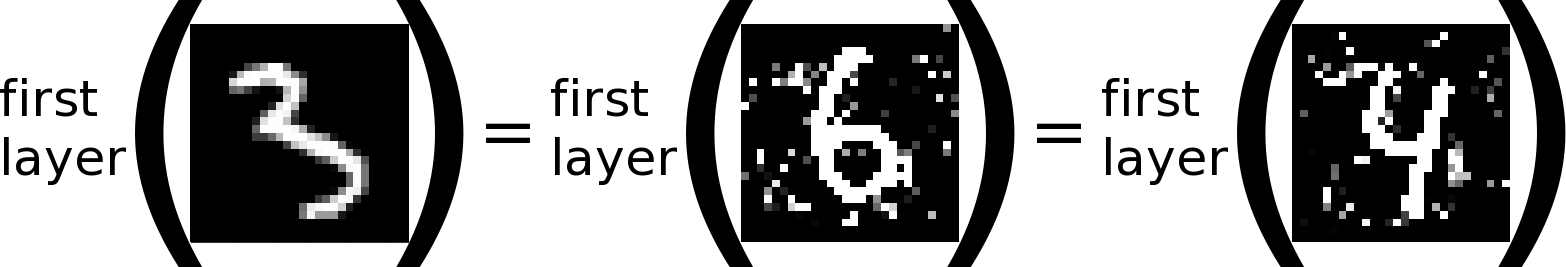}
\hfill
\includegraphics[width=0.485\textwidth]{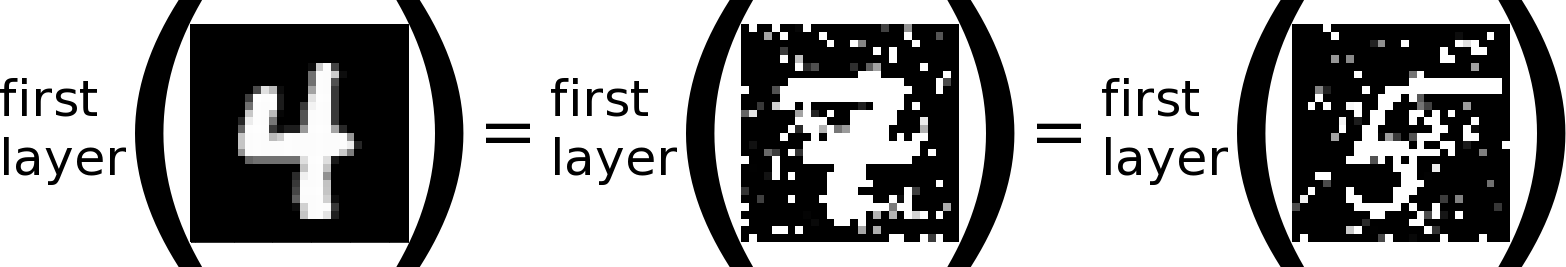}
\caption{Invariances of the first layer (100 $\relu$ neurons) of a vanilla multilayer perceptron (MLP). Despite the semantically very different examples, the features are identical as the original image "3" and the two perturbed variants "6" and "4" are in the same preimage. Further details in Appendix~\ref{app:invarianceMNIST}.}
\label{opener_image}
\end{figure}

While there has been a growing effort to rigorously study the mathematics behind deep neural networks (DNNs), see \citep{mathOverview} for an overview, their inner workings are only sparsely understood so far. Although many open questions have been tackled -- like the expressive power of neural networks \citep{express}, the matter of generalization \citep{generalization} and the difficulties of optimizing a non-convex loss landscape \citep{lossSurface} -- the inner workings of a particular network is often at best partially understood. Besides a given model's performance on the test set the interpretation of its decisions \citep{MonDSP18} and the analysis of its sensitivity to adversarial perturbations \citep{Szegedy} are key steps to assess the nature of its behavor. \\
Despite of, or maybe due to, the relevancy of adversarial examples,
it is of similar importance to better our understanding of the opposite effect: Which perturbations $\Delta x$ do not (or only little) affect the outcome of the network $F$? This question can be addressed for a given input datapoint $x$ via searching for all $\Delta x$, such that
\begin{align*}
F(x) = F(x+ \Delta x) \quad \text{(invariant) or} \quad \|F(x) - F(x + \Delta x)\| \leq \varepsilon \quad \text{(robust)},
\end{align*}
where a small $\varepsilon > 0$ is given. These properties can be crucial for many discriminative tasks in order to contract the space along uninformative directions \citep{mallat}. However, a model would be flawed if perturbations that alter the semantics only have a minor impact on the features of the network. In Figure \ref{opener_image} such examples are shown, where perturbations entirely change its semantics while all the network's features -- even the first layer's -- are exactly the same for the original and perturbed image. \\
A natural way to address these properties is by studying the invertibility: If $F$ is invariant to perturbations $\Delta x$, then $x$ and $\Delta x$ lie in the preimage of the output $z = F(x)$ i.e.~$F$ is not uniquely invertible. Robustness towards large perturbations induces an instable inverse mapping as small changes in the output can be due to large changes in the input. Most notably, this analysis is dual to adversarial examples \citep{Szegedy} where small perturbations induce large changes in the outcome, which shows instabilities in the forward mapping. \\
In this work we address the issue of invariance via analyzing preimages and studying the stability of the inverse mapping. Besides the importance of understanding the behavior of a model, further applications of studying the inverse include: input reconstruction \citep{Mahendran2015} \citep{Mahendran16} to backtrack properties from feature space to input space or inverse problems with learned forward models \citep{Jensen} \citep{Liang}. As the approximation capabilities of neural networks were recently even used to learn solutions of partial differential equations \citep{Sirignano}, parameter estimation problems often formulated as inverse problems would require the inversion of networks.

\subsection{Related Work and Contributions}
While analyzing invariance and robustness properties is a major topic in theoretical treatments of deep networks \citep{mallat}, studying it via the inverse is less common. Several works like \citep{Mahendran2015} or \citep{Dosovitskiy} focus on reconstructing inputs from features of convolutional neural networks (CNNs) to visualize the information content of features. Instead, we investigate potential mechanisms affecting the invertibility. \citep{preimage} gives a first geometrical view on the shape of preimages of outputs from $\relu$ layers, which is directly related to the question of injectivity of the mapping under $\relu$. \citep{crelu} analyzes the reconstruction property of cReLU (concatenated $\relu$); however, the more general situation of using the standard rectifier is not studied. A notable other line of work assumes random weights in order to derive guarantees for invertibility, see \citep{gilbert} or \citep{Arora}, whereas we aim to theoretically derive means to uncover mechanisms of rectifier networks without assumptions on the weights. \\
Moreover, several reversible network structures were recently proposed \citep{revnet} \citep{changRevNet} \citep{iRevnet}. Most notably, in \citep{iRevnet} a bijective network, up to its last layer, was trained successfully on ImageNet which does not exhibit any invariance. However, the network is very robust towards many directions in the input which is reflected in a strongly instable inverse. Hence, even carefully designed network show at least one of the two effects (invariance and robustness) studied in this work. Especially stability has seen growing interest due to adversarial examples \citep{Szegedy}, yet stability is mostly studied with respect to the forward mapping, see e.g. \citep{parseval}. \\
Two main resources for our view of rectifier networks as piecewise linear models are \citep{montufar} and \citep{express}. Closest to our approach is the work of \citep{Bruna} on global statements of injectivity and stability of a single layer including $\relu$ and pooling. The authors focus on global injectivity and stability bounds via combinatorial statements over all configurations attainable by $\relu$ and pooling. These conditions are valid on the entire input space, while the restriction to parts of the input space may be far from these worst-case conditions. \\
Our contributions are as follows:
\begin{itemize}
\item We provide computable conditions when the preimage of an output point for one $\relu$-layer is finite, infinite or a single point (injective). In contrast to \citep{Bruna}, where conditions are given for all input points, we refine this locally for a single point. Afterwards we numerically investigate the properties of preimages for a trained network. (See Section~\ref{sec:uniquenss}.)
\item We study the stability of the inverse via analyzing the linearization at a point in input space, which is accurate within a polytope. We provide upper bounds on the smallest singular value and prove how the removal of uncorrelated features could effect the stability of the inverse mapping. Furthermore, we provide numerical results showing the (in-)stability of the inverse of rectifier networks. (See Section~\ref{sec:stability}.)
\end{itemize}

\subsection{Notation}
\label{sec:notation}
\begin{itemize}
\begin{minipage}{0.45\linewidth}
\item Input: $x=x^0\in \mathbb{R}^{d_0}$, sometimes shortened to $d:=d_0$.
\item Pre-activations: $z^l=A^lx^{l-1}+b^l\in\mathbb{R}^{d_l},$ with weight matrix $A^l\in\mathbb{R}^{d_{l}\times d_{l-1}}$ and bias $b^l\in\mathbb{R}^{d_l}$.
\end{minipage}
\hfill
\begin{minipage}{0.45\linewidth}
\item Activation: $x^l = \phi(z^l)\in\mathbb{R}^{d_l},$ where $\phi:\mathbb{R}\to\mathbb{R}$ the pointwise applied activation function, if not specified differently $g:=\relu$.
\item Number of layers: $L\in\mathbb{N}$
\end{minipage}
\item Entire network: $F:\mathbb{R}^d\ni x\mapsto F(x):=z:=z^L\in\mathbb{R}^{d_L}$, sometimes short $D:=d_L$.
\end{itemize}
For matrices $A\in\mathbb{R}^{m\times n}$ and $I\subset[m] := \{1, \dots, m\}$, $A|_I$ denotes the matrix consisting of the rows of $A$ whose index is in set $I$ -- analogously for vectors. Also $A|_{y>0}$ describes the restriction to the index set $\{i:y_i>0\}$ for $y\in\mathbb{R}^m$, analogously for $<,=,\le,\ge$. For vectors $y\in\mathbb{R}^m$, $y > 0$ is the elementwise relation, analogously for $<,=,\le,\ge$. Furthermore, we define $ \mathcal{N}(A) $ as the null space of the matrix $ A. $ \\
For every matrix $A\in\mathbb{R}^{m\times n}$ with the rows $a_i$, $i \in [m]$, we associate the set $A=\{a_i\}_{i=1}^m$. Vice versa, we associate every finite set in $\mathbb{R}^n$ with a matrix (only possible up to permutation of the indices).

\section{Preimages of ReLU Layer}
\label{sec:uniquenss}
\subsection{Theoretical Analysis}
In this section, we analyze different kinds of preimages of a $ \relu $-layer and investigate under which conditions the inverse image of a given point is a singleton (a set containing exactly one element) or has finite/infinite volume. 
These conditions will yield a simple algorithm able to distinguish between these different preimages, which is applied in Section \ref{sec:retrievalnumerics}.\newline
For the analysis of pre-images of a given output one can study single layers separately or multiple layers at once. However, since the concatenation of two injective functions is again injective while a non-injective function followed by an injective function is non-injective, studying single layers is crucial. We therefore develop a theory for the case of single layers in this section.
Notice that in case of multiple layers one is also required to investigate the image space of the previous layer.\\
We will focus our study on the most common activation function, $\relu$. One of its key features is the non-injectivity, caused by the constant mapping on the negative half space. It provides neural networks with an efficient way to deploy invariances. Basically all other common activation functions are injective, which would lead to a straightforward analysis of the preimages. However, injective activations like $\mbox{ELU}$~\cite{clevert2015fast} and $\mbox{Leaky ReLU}$~\cite{maas2013rectifier} only swap the invariance for robustness, which in turn leads to the problem of having instable inverses. This question of stability will be analyzed in more detail in Section \ref{sec:stability}. \newline
We start by introducing one of our main tools -- namely the \textit{omnidirectionality}.
\begin{definition}[Omnidirectionality]\mbox{}
\label{def:omni}
\begin{enumerate}[\bfseries i)]
\item $A\in\mathbb{R}^{m\times n}$ is called omnidirectional if there exists a unique $ x\in \mathbb{R}^n,$ such that $ Ax\leq 0$ (component-wise).
\item $A\in\mathbb{R}^{m\times n}$ and $b\in\mathbb{R}^m$ are called omnidirectional for the point $p\in\mathbb{R}^n$ if $A$ is omnidirectional and $b=-Ap.$
\end{enumerate}
\end{definition}

\begin{corollary} \label{cor_omni_short}
The following statements are equivalent:
\begin{enumerate}[\bfseries i)]
\item $A \in \mathbb{R}^{m \times n}$ is omnidirectional.
\item Every linear open halfspace in $\mathbb{R}^n$ contains a row of $A$. \label{cor:omni:Hyperplanes}
\item $Ax\le0$ implies $x=0$, where $x \in \mathbb{R}^n$.
\end{enumerate}
\end{corollary}
\begin{figure}
\center
\resizebox{0.2\textwidth}{!}{
\begin{tikzpicture}
\draw[|->, thick, black](0,0) -- (0,1);
\draw[|->, thick, black](0,0) -- (-0.8660254, -0.5);
\draw[|->, thick, black](0,0) -- (0.8660254, -0.5);

\draw[-, thin, gray](-2,0) -- (2,0);
\draw[-, thin, gray](1.5*-0.5,1.5*0.8660254) -- (1.5*0.5,1.5*-0.8660254);
\draw[-, thin, gray](-1.5*0.5,-1.5*0.8660254) -- (1.5*0.5,1.5*0.8660254);

\draw[dotted] (0,0) .. controls (1*0.8,0.5*0.8) .. (1.5*0.8,1*0.8) node[pos=1]{p};
\end{tikzpicture}
}\hspace*{0cm}
\resizebox{0.2\textwidth}{!}{
\begin{tikzpicture}
\draw[|->, thick, black](0,0) -- (0,1);
\draw[|->, thick, black](0,0) -- (-0.8660254, 0.5);
\draw[|->, thick, black](0,0) -- (0.8660254, 0.5);

\draw[-, thin, gray](-2,0) -- (2,0);
\draw[-, thin, gray](-1.5*0.5,-1.5*0.8660254) -- (1.5*0.5,1.5*0.8660254);
\draw[-, thin, gray](1.5*0.5,-1.5*0.8660254) -- (-1.5*0.5,1.5*0.8660254);

\end{tikzpicture}
}\hspace*{0cm}
\resizebox{0.2\textwidth}{!}{
\begin{tikzpicture}
\draw[|->, thick, black](0,0) -- (0,1);
\draw[|->, thick, black](0,0) -- (-0.8660254, 0.5);
\draw[|->, thick, black](0,0) -- (0.8660254, 0.5);
\draw[|->, thick, black](0,0) -- (1,0);

\draw[-, thin, gray](-2,0) -- (2,0);
\draw[-, thin, gray](-1.5*0.5,-1.5*0.8660254) -- (1.5*0.5,1.5*0.8660254);
\draw[-, thin, gray](1.5*0.5,-1.5*0.8660254) -- (-1.5*0.5,1.5*0.8660254);
\draw[-, thin, gray](0,-1.3) -- (0,1.3);
\end{tikzpicture}
}\hspace*{0cm}
\resizebox{0.2\textwidth}{!}{
\begin{tikzpicture}
\draw[|->, thick, black](0,0+0.5) -- (0,1+0.5);
\draw[|->, thick, black](0-0.5,0) -- (-0.8660254-0.5, -0.5);
\draw[|->, thick, black](0,0) -- (0.8660254, -0.5);

\draw[-, thin, gray](-2,0+0.5) -- (2,0+0.5);
\draw[-, thin, gray](1.5*-0.5-0.33,1.5*0.8660254-0.3) -- (1.5*0.5-0.8,1.5*-0.8660254+0.5);
\draw[-, thin, gray](-1.5*0.5+0.29,-1.5*0.8660254+0.5) -- (1.5*0.5-0.19,1.5*0.8660254-0.3);

\end{tikzpicture}
}
\caption{Gray lines are hyperplanes with normal vectors (arrows) from the rows of $A$ and translation $b$. Left: Omnidirectional tuple $(A,b)$ for $p\in\mathbb{R}^2$, as hyperplanes intersect in $p$ and normal vectors are omnidirectional. Two in the middle: Intersection in $p$, but vector-free halfspaces (hence, not omnidirectional). Right: hyperplanes do not intersect in a point, but normal vectors are omnidirectional.}
\label{omni_and_not_omni_illustration_for_point}
\end{figure}
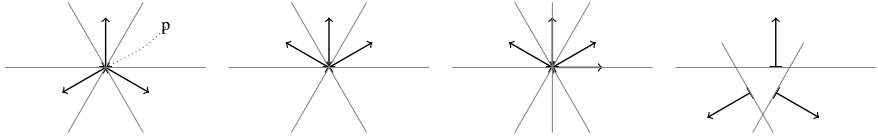

Thus, for every direction of a hyperplane through the origin forming two halfspaces, there is a vector from the rows of $A$ inside each open halfspace, hence the term \textit{omnidirectional} (see Figure~\ref{omni_and_not_omni_illustration_for_point} for an illustration). Note that the hyperplanes are due to $\relu$ as it maps the open halfspace to positive values and the closed halfspace to zero. A straightforward way to construct an omnidirectional matrix is by taking a matrix whose rows form a spanning set $\mathcal{F}$ and use the vertical concatenation of $\mathcal{F}$ and $-\mathcal{F}$. This idea is realated to $\mbox{cReLU}$~\cite{crelu}.\newline
More importantly, omnidirectionality is directly related to the $ \relu $-layer preimages and will provide us with a computable method to characterize their volume (see Theorem \ref{theorem_pre-images}). To analyze such inverse images, we consider
\begin{equation} \label{equ_relu_layer}
y=\relu(Ax+b)
\end{equation}
for a given output $y\in\mathbb{R}^m$ with $A\in\mathbb{R}^{m\times n}$, $b\in\mathbb{R}^m$ and $x\in\mathbb{R}^n.$
If we know $A, b$ and $y$, we can write \eqref{equ_relu_layer} as the following mixed linear system:
\begin{align}
\label{mixed_system_equality}
A|_{y>0}x+b|_{y>0}&=y|_{y>0}\\
\label{mixed_system_inequality}
A|_{y=0}x+b|_{y=0}&\le0.
\end{align}
\vspace{-5mm}
\begin{remark} It is possible to enrich the mixed system to include conditions/priors on $x$ (e.g.~$x\in\mathbb{R}^n_{\ge0}$). \end{remark}
The inequality system in \eqref{mixed_system_inequality} links its set of solutions and therefore the volume of the preimages of the $ \relu $-layer with the omnidirectionality of $ A $ and $ b. $ Defining $\overline A\coloneqq AO^T, $ where $O\in\mathbb{R}^{k\times n}$ denotes an orthonormal basis of $\sN(A|_{y>0})$ with $k:=\dim \sN(A|_{y>0})$ and $\overline b := b|_{y\le0} + A|_{y\le0}(P_{\sN(A|_{y>0})^\perp}x)$ leads to the following main theorem of this section, which is proven in Appendix \ref{app:uniqueness}.
\begin{theorem}[Preimages of $ \relu $- layer] \label{theorem_pre-images}
	The preimage of a point under a $\relu$-layer is a
	\begin{enumerate}[\bfseries i)]
		\item singleton, if and only if there exists an index set $ I $ for the rows of $ \overline A $ and $\overline b, $ such that $ (\overline{A}|_I, \overline{b}|_I) $ is omnidirectional for some point $ p\in \mathbb{R}^n. $ \label{theorem_pre-images_singleton}
		\item compact polytope with finite volume, if and only if $\overline A$ is omnidirectional. \label{theorem_pre-images_compact}
	\end{enumerate}
\end{theorem}
Thus, omnidirectionality allows in theory to distinguish whether the inverse image of a $ \relu $-layer is a singleton, a compact polytope or has infinite volume. However, obtaining a computable method to decide whether a given matrix is omnidirectional is crucial for later numerical investigations. For this reason, we will go back to the geometrical perspective of omnidirectionality (see Figure~\ref{omni_and_not_omni_illustration_for_point}). This will also help us to get a better intuition on the frequency of occurence of the different preimages. The following Theorem \ref{thm_convex_hull} gives another geometrical interpretation of omnidirectionality besides Corollary \ref{cor_omni_short}\ref{cor:omni:Hyperplanes}, whose short proof is given in Appendix \ref{app:uniqueness}.
\begin{theorem}[Convex hull] \label{thm_convex_hull}
A matrix $A\in\mathbb{R}^{m\times n}$ is omnidirectional if and only if $0\in\interior{\Conv(A)},$ where $\interior{\Conv(A)}$ is the interior of the convex hull spanned by the rows of $A$ (see Definition~\ref{def_convex_hull} in Appendix \ref{app:uniqueness}). 
\end{theorem}

Therefore, the matrix must contain a simplex in order to be omnidirectional, as the convex hull of the matrix $A\in \mathbb{R}^{m\times n}$ has to have an interior. Hence, we have the following:
\begin{corollary} \label{corollary_omni_set_lines}
	If $ A\in \mathbb{R}^{m\times n} $ is omnidirectional, then $m>n.$
\end{corollary}
Considering the geometric perspective that the tuple $(A\in\mathbb{R}^{m\times n}, b\in\mathbb{R}^m)$ is omnidirectional for a point $p\in \mathbb{R}^n,$ if and only if the $m$ hyperplanes generated by the rows of $A$ with bias $b$ intersect at $p$ and their normal vectors (rows of $A$) form an omnidirectional set. We can use Corollary \ref{corollary_omni_set_lines} to conclude that singleton preimages of $\relu $-layers are very unlikely to happen in practice (if we do not design for it), since a necessary condition is that $n+1$ hyperplanes have to intersect in one point in $\mathbb{R}^n.$ Therefore we conclude, that singleton preimages of $\relu$ layers in practice only and exclusively occurs, if the mixed linear system already has sufficient linear equalities.\newline
Furthermore, the above results can be used to derive an algorithm to check whether a preimage of a given output is finite, infinite or just a singleton. A singleton inverse image is obtained as long as $ \rank(A|_{y>0})=n $ holds true, which can be easily computed. To distinguish preimages with finite and infinite volumes, it is enough to check if $ \overline{A} $ is omnidirectional (see Theorem \ref{theorem_pre-images}\ref{theorem_pre-images_compact}), which can be done numerically by using the definition of the convex hull, Theorem \ref{thm_convex_hull} and Corollary \ref{corollary_omni_set_lines}. This leads to a \textit{linear programming problem}, which is presented in Appendix \ref{app:algoRetrieval}.

\subsection{Numerical Analysis}
\label{sec:retrievalnumerics}
In this subsection we demonstrate for a simple model that the preimage of a layer can be a singleton, infinite or finite depending on the given point. For this purpose, we trained a MLP with two hidden $\relu$ layers of size $3500$ and $784$ and a 10 neuron softmax output layer on MNIST~\cite{mnist}. We chose the layer size of $3500$, because the likelihood of having roughly $784$ (input dimension of MNIST) positive outputs was high for this setting.
In Figure \ref{plot_numeric_finite_preimage}, we plotted how many samples of the test set have infinite (red curve) or finite (blue curve) preimages over the number of positive outputs. It can be assumed that all samples which have more or equal to 784 (the input dimension) positive outputs have a singleton preimage and are therefore finite. In the dark gray region between $723$ and $784$, both effects occurred, which can be seen by the overlap of the red and blue curve. \newline
To determine whether a preimage for less than 784 positive outputs was compact we used Theorem \ref{theorem_pre-images}\ref{theorem_pre-images_compact} and the algorithm described in Appendix \ref{app:algoRetrieval}.\newline
We also conducted first experiments whether the compactness of the pre-image has an influence on the range of manipulations à la Figure~\ref{opener_image}. However, compact pre-images only occurred in cases where the number of positive outputs resulted in a linear system which already determined the image up to a level where manipulations were imperceptible. A thorough study is left for future work.

\begin{figure}
	\centering
	\begin{minipage}{0.55\textwidth}
		\center
	\subfile{figures/removal_of_vectors}
	\end{minipage}
	\hfill
	\begin{minipage}{0.43\textwidth}
		\center
		\subfile{tikz_images/plot_numeric_finite_preimage}
	\end{minipage}
\end{figure}

\section{Stability}
\label{sec:stability}
\subsection{Theoretical Analysis}
In this section we analyze the robustness of rectifier MLPs against large perturbations via studying the stability of the inverse mapping. Concretely, we study the effect of $\relu$ on the singular values of the linearization of network $F$. While the linearization of a network $F$ at some point $x$ only provides a first impression on its global stability properties, the linearization of $\relu$ networks is exact in some neighborhood due to its piecewise-linear nature \citep{express}. In particular, the input space $\mathbb{R}^d$ of a rectifier network $F$ is partitioned into convex polytopes $P_F$, corresponding to a different linear function on each region. Hence, for each polytope $P$ in the set of all input polytopes $P_F$, the network $F$ can be simplified as $F(x) = A_P x + b_P$ for all $x \in P$. \\
Additionally, each of these matrices $A_P$ can be written via a chain of weight matrix multiplications incorporating the effect of $\relu$, see \citep{dataJacobian}. In particular let $D_I$ denote a diagonal matrix with $D_ {ii}=1$ for $i \not \in I$ and $D_{ii}=0$ for  $i \in I$. Then $A_P$ of a network with $L$ layers can be written as
\begin{align*}
A_P = A^L D_{I^{L-1}} A^{L-1} \cdots D_{I^1} A^1,
\end{align*}
where $A^l$ are the weight matrices of layer $l$ and $I^{l} := \{i \in [d_l]: (A^l x^{l-1} + b^l)_i \leq 0\}$. Thus, the effect of $\relu$ is incorporated in the diagonal matrices $D_{I^l}$ which set the rows of $A^l$ with indices from $I^l$ to zero. For the purpose of studying all possible local behaviors, we define admissible index sets $I^l$ following, with slight modifications, \citep{Bruna}:
\begin{definition}(Admissible index sets)\\
An index set $I^l$ for a layer $l$ is admissible if 
\begin{align*}
\bigcap_{i \not \in I^l} \{x^l: \langle x^l, a^l_i\rangle > - b_i\} \cap \bigcap_{i \in I^l} \{x^l: \langle x^l, a^l_i\rangle \leq -b_i\} \neq \emptyset.
\end{align*}
\end{definition}
Of special interest for such an analysis is the range of possible effects by the application of the rectifier. Since the effect by $\relu$ corresponds to the application of $D_I$ for admissible $I$, we now turn to studying the changes of the singular values due to $D_I$. By considering the matrix $A$, e.g. representing the chain of matrix products up to layer $l$, the effect of $\relu$ can be globally upper bounded:
\begin{lemma}(Global upper bound for largest and smallest singular value)\\
\label{lem:globalBound}
Let $\sigma_l$ be the singular values of $D_I A$. Then for all admissible index sets $I$, the smallest non-zero singular value is upper bounded by
\begin{align*}
\min \{\sigma_l:  \sigma_l >0 \} \leq \tilde{\sigma}_k,
\end{align*}
where $k = N - |I|$ and $\tilde{\sigma}_1 \geq ... \geq \tilde{\sigma}_N > 0$ are the non-zero singular values of $A$. \\
Furthermore, the largest singular value is upper bounded by
\begin{align*}
\max \{\sigma_l:  \sigma_l >0 \} \leq \tilde{\sigma}_1.
\end{align*}
\end{lemma}
Lemma \ref{lem:globalBound} analyzes the best case scenario with respect to the highest value of the smallest singular value. While this would yield a more stable inverse mapping, one needs to keep in mind that the nullspace $\sN(A_P)$ grows by the corresponding elimination of rows via $D_I$. Moreover, reaching this bound is very unlikely as it requires the singular vectors to perfectly align with the directions that collapse due to $D_i$. Thus, we now turn to study effects which could happen locally for some input polytopes $P$. \\
An example of a drastic effect through the application of $\relu$ is depicted in Figure \ref{fig:corrFeat}. Since one vector is only weakly correlated to the removed vector and the situation is overdetermined, removing this feature for some inputs $x$ in the blue area leaves over the strongly correlated features. While the two singular values of the 3-vectors-system were close to one, the singular vectors after the removal by $\relu$ are badly ill-conditioned. As many modern deep networks increase the dimension in the first layers, redundant situations as in Figure \ref{fig:corrFeat} are common, which are inherently vulnerable to such phenomena. For example, \citep{uncorrFeat} proposes a regularizer to avoid such strongly correlated features. The following lemma formalizes the situation exemplified before:
\begin{lemma}(Removal of weakly correlated rows)\\
\label{lem:corrRows}
Let $A\in \mathbb{R}^{m \times n}$ with rows $a_j$ and $I\subseteq[m]$. For a fixed $k \in I$ let $a_k \in \sN(D_I A)^\bot$. Moreover, let
\begin{align}
\label{eq:corrCond}
\forall j \not \in I: |\langle a_j, a_k\rangle| \leq c \frac{\|a_k\|_2}{\sqrt{M}},
\end{align}
with $M = m - |I|$ and constant $c>0$. Then for the singular values $\sigma_l \neq 0$ of $D_I A$ it holds
\begin{align*}
0 < \sigma_K = \min \{\sigma_l: \sigma_l \neq 0\} \leq c.
\end{align*}
\end{lemma}
(Note that $I$ has to be admissible when considering the effect of $\relu$.)

Lemma \ref{lem:corrRows} provides an upper bound on the smallest singular value, given a condition on the correlation of all $a_j$ and $a_k$. However, the condition \eqref{eq:corrCond} depends on the number $N$ of remaining rows $a_j$. Hence, in a highly redundant setting even after removal by $\relu$ (large $N$), $c$ needs to be large such that the correlation fulfills the condition. Yet, in this case the upper bound on the smallest singular value, given by $c$, is high. We discuss this effect further in the Appendix \ref{app:corrAnalysis}.\\
We now turn our attention to the effect of multiple layers and ask whether the use of multiple layers results in a different situation than a 1-hidden layer MLP. In particular: Can the application of another layer have a pre-conditioning effect yielding a stable inverse? What happens when we only compose orthogonal matrices which have stable inverses? Note that a way to enforce an approximate orthogonality constraint was proposed for CNNs in \citep{parseval}, however only for the filters of the convolution. \\
For both situations the answer is similar: the nonlinear nature of $\relu$ induces locally different effects. Loosely speaking, we apply the layer $A^l$ to different linearizations depending on the input region $P$. Thus, if we choose a pre-conditioner $A^l$ for a specific matrix $A^{l-1}_P$, it might not stabilize the matrix product for matrices $A^{l-1}_{P^*}$ corresponding to different input polytopes $P^*$. For the case of composing only orthogonal matrices, consider a network up to layer $l-1$ where the linearization $A^{l-1}_P$ has orthogonal columns (assume the network gets larger, thus $A^{l-1}_P$ has more rows than columns). Then, the application of $\relu$ in form of 
\begin{align}
\label{eq:orthProd}
A^l D_{I^l} A^{l-1}_P
\end{align} 
removes the orthogonality property of the rows of $A^{l-1}_P$, if setting entries in the rows from $I^l$ to zero results in non-orthogonal columns. This effect is likely, especially when considering dense matrices, hence $D_{I^l} A^{l-1}_P$ is for some $I^l$ not orthogonal. Thus, the matrix product \ref{eq:orthProd} is not orthogonal, resulting in decaying singular values. \\
This is why, even when especially designing the network by e.g. orthogonal matrices, stability issues with respect to the inverse arise. To conclude this section, we remark that the presented results are rather of a \textit{qualitative} nature showcasing effects of $\relu$ on the singular values. Yet, the analysis does not require any assumptions and is thus valid for any MLP (including CNNs without pooling). To give an idea of \textit{quantitative} effects we study numerical examples in the subsequent subsection.

\subsection{Numerical Analysis}
\label{sec:numericsstability}
\begin{figure}
	\centering
	\begin{minipage}{.41\textwidth}
		\centering
		\vspace{-5mm}
		\includegraphics[width=1.0\textwidth]{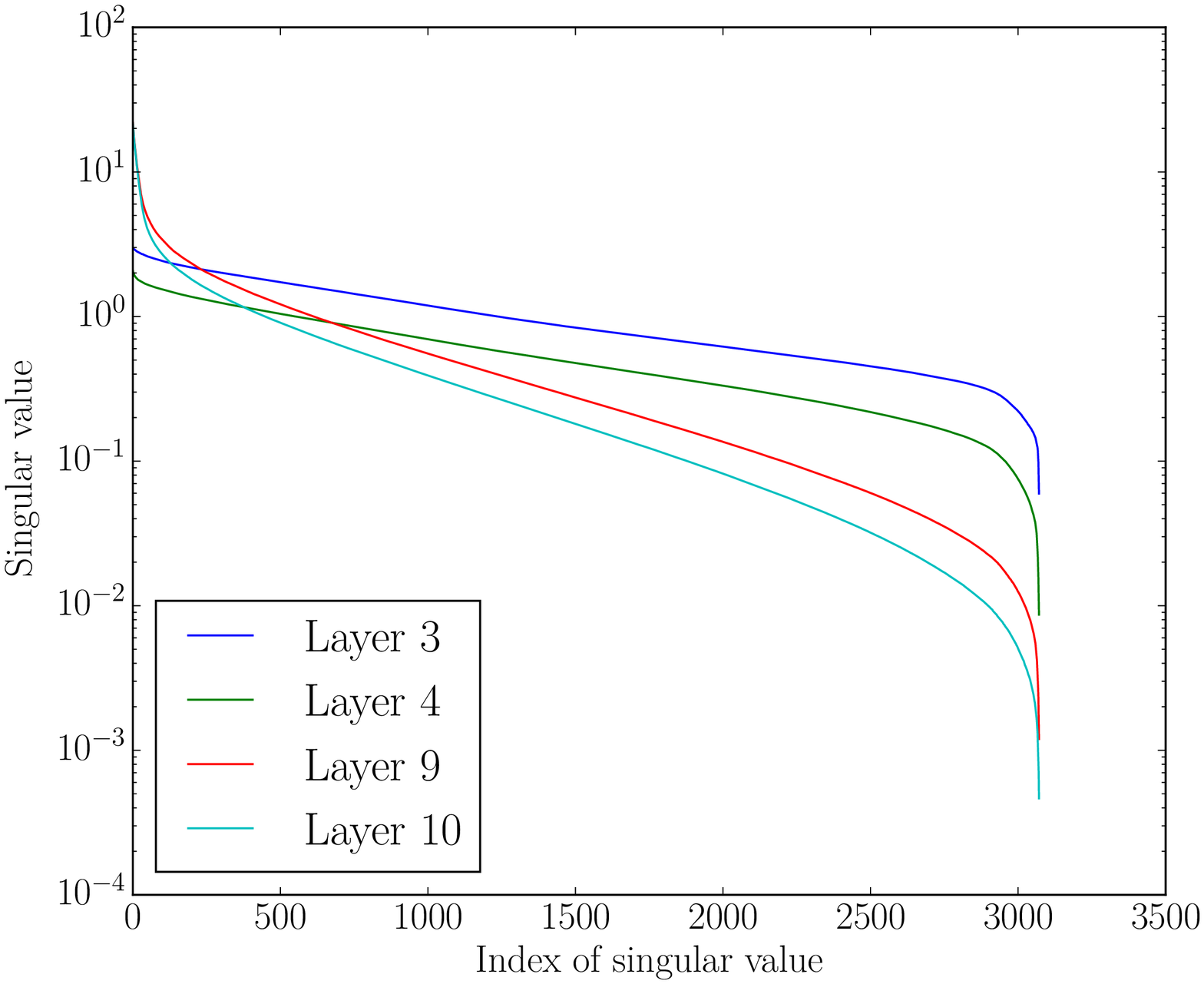} 
		\caption{Effect of $\relu$ on the singular values for WideCifar. The curves show the effect in layer 2 (layer 3 and 4 in the legend, because $\relu$ is counted as an extra activation layer) and layer 5 (layer 9 and 10), where each curve is the median over 50 samples.}
		\label{fig:effectRelu}
	\end{minipage}%
	\hfill
	\begin{minipage}{0.41\textwidth}
		\vspace{-13mm}
		\centering
		\includegraphics[width=1.0\textwidth]{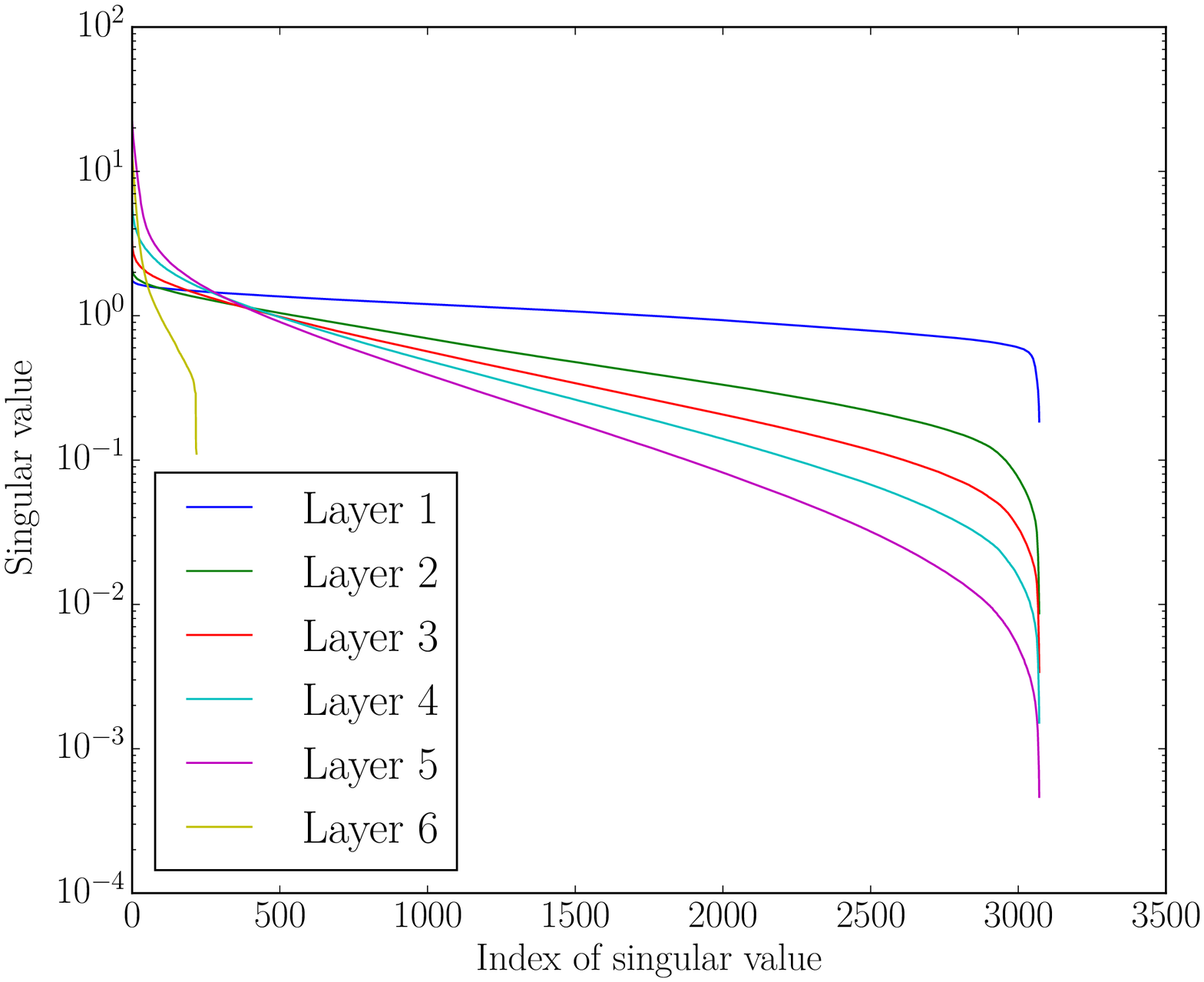} 
		\caption{Decay of singular values over the layers of the network. Here, each layer includes the convolution and $\relu$-activation. Reported number are taken from median over 50 samples. }
		\label{fig:singDecayLayers}
	\end{minipage}
\end{figure}

\begin{figure}
	\vspace{-9mm}
		\centering
		\resizebox{0.46\textwidth}{!}{\input{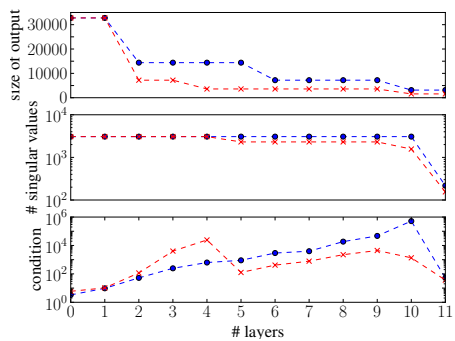}}
		\caption{In blue are the results from WideCIFAR and in red ThinCIFAR. Top: number of output units per layer, Middle: number of singular values, Bottom: Behavior of condition number, each curve over the layers. Here, layers are split into conv-layer and $\relu$-activation layer. Singular values and condition number are the median over 50 samples from the CIFAR10 test set.}
		\label{fig:condNumSing}
\end{figure}

In this section we conduct simple experiments on CIFAR10 \cite{krizhevsky2009learning} to numerically study the stability of CNNs. We simplify the CNNs by using only strides instead of pooling, and furthermore use no residual connections and batch-normalization layers. Thus, the architectures fit to the theoretical study as the strided discrete convolution can be written as a matrix-vector multiplication. Details on the used CNNs and training setup are in Appendix \ref{app:architectures}. Furthermore, we remark that the linearization of a network $F$ for an input point $x^0$ can be computed via backpropagation.\\
  Using the linearization obtained by backpropagation, we proceed by computing the SVD. In particular, we are interested in the entire distribution of singular values. Computing all singular values is numerically expensive and scales with the input and output size of the linearized network. Especially, early CNN-layers have high dimensional outputs which may cause memory issues when computing the entire SVD. We thus choose a small CNN trained on CIFAR10 as these inputs are only of size $32\times 32 \times 3$. In order to scale this analysis up to e.g. ImageNet with VGG-networks, a restriction to a window of the input image is necessary in order to compute the full SVD of early layers. See \citep{iRevnet}, where the singular values restricted to input windows were used to estimate the stability of the entire i-RevNet trained on ImageNet. \\
  \\
  \textbf{Effect of $\relu$:} We start the evaluation by looking into the effect of $\relu$ on the singular values of a layer. In the theoretical section, Lemma \ref{lem:corrRows} considered the case of removal of weakly correlated rows. In these situations, $\relu$ could remove rows of the weight matrices resulting in drastically smaller singular values after $\relu$. However, as Figure \ref{fig:effectRelu} shows for the layers 2 and 5 of WideCIFAR, this effect is not present in this example. In this figure, the decay of the median singular values is shown using 50 samples from the CIFAR10 test set propagated until layer 2 and 5. The blue and red lines correspond to the linearization without $\relu$-activation, while the green and turquoise lines are singular values after $\relu$. While $\relu$ results in smaller singular values, as discussed in the upper bounds in Lemma \ref{lem:globalBound}, they are only slightly smaller. To better understand this effect, we numerically compute the conditions from Lemma \ref{lem:corrRows} and discuss the implications in Appendix \ref{app:corrAnalysis}. In summary from the results in the appendix, the above example is too redundant to exhibit strong effects due to removal of rows. \\
  \\
  \textbf{Effect of multiple layers:} We now move to the development of the singular values over several layers. Figure \ref{fig:singDecayLayers} shows the decay in convolutional layers (layers 1-6, after application of $\relu$). While the shape of the curve is similar for layer 1-5, it can be seen that the largest singular value grows, while the small singular values decrease significantly. Note that this growth of the largest singular values is in line with observations for adversarial examples, see \citep{Szegedy}. While many defense strategies like \citep{parseval} or \citep{singValBound} focus on the largest singular value, the behavior of the smaller singular values are of special interest for the stability of the inverse. Another effect to note, is the cut-off curve for layer 6 in yellow. Here there are way fewer singular values than in previous layers. This is caused by the downsampling in this layer by strided convolution and the row removal by $\relu$. \\
  The same experiments were conducted for ThinCIFAR and for the MLP from Section \ref{sec:retrievalnumerics}, see Appendix \ref{app:furtherExp}.\\
  \\
  \textbf{Trade-off between stability and invariance:} In the example above, it could be seen that the inverse for layer 6 would be more stable, while there are many invariance directions (zero singular value). To further investigate this trade-off, Figure \ref{fig:condNumSing} compares the output size, the condition number and the number of non-zero singular values vs.~the layers for WideCIFAR and ThinCIFAR. In combination with lower output dimension, $\relu$ has a different effect for ThinCIFAR. The number of singular values decreases in layer 5, which cuts off the smallest singular values, resulting in a lower condition number. On the other hand, there are more directions where the network is invariant within the corresponding linear region. \\
In conclusion, these experiments showed how $\relu$ affects the singular value decay and exposed a trade-off between stability and information loss. This study investigated only small networks on MNIST and CIFAR10, in order to reduce computational costs. However, a study on larger networks and datasets would be of great interest and is left for future work.


\section{Conclusion and Outlook}
We presented a approach to better understand the invariance and robustness properties of $\relu$
networks via studying its inverse. Our analysis yielded computable conditions under which the preimage of a $\relu$ layer is a point, finite or infinite. We showed how to analyze the inverse stability using the singular values of the linearization. This view provided insights into mechanisms that effect stability and allowed simple experiments to study properties of the inverse of CNNs. \\
Most noticeable, our analysis lays a well founded starting point for further studies on robustness and invariance properties. In future work we plan to further investigate methods to construct examples from the preimage of features deeper in the network (invariance) or finding large perturbations which do little effect (robustness). Furthermore, a more accurate analysis of the inverse stability should incorporate nonlinear effects like moving between linear regions of rectifier networks. In addition, our study considered general weight matrices, whereas a restriction to CNNs might enable a tighter theoretical analysis. Finally, studying the invertibility could be of major interest in the context of inverse problems with forward models learned by neural networks.

\section{Acknowledgments}
We thank Emily King and Christian Etmann for helpful discussions and feedback on the draft and Dmitry Feichtner-Kozlov for pointing out Stiemke's theorem. Furthermore, we gratefully acknowledge the financial support from the German Science Foundation for RTG 2224 "$\pi^3$: Parameter Identification - Analysis, Algorithms, Applications". 

\bibliography{main_lit}
\bibliographystyle{abbrvnat}

\newpage
\onecolumn
\setcounter{section}{0}
\renewcommand*{\thesection}{A\arabic{section}}
\section{Appendix for Section~\ref{sec:uniquenss}}
\label{app:uniqueness}

\begin{corollary}
\label{corollary_omnidirectional}
For $A\in\mathbb{R}^{m\times n}$ the following statements are equivalent.
\begin{enumerate}
\item $A$ is omnidirectional.
\item $\nexists x\ne0:Ax\le0$
\item $\nexists x\gneq0:Ax\le0$
\item Every linear open halfspace contains a row of $A$.
\item $Ax\le0$ implies $x=0$.
\end{enumerate}
Here $z\lneq0$ means $z\in\mathbb{R}^n_{\le0}$ and $z\ne0$.
\end{corollary}

\begin{definition} (Convex hull)\\
\label{def_convex_hull}
For $A\in\mathbb{R}^{m\times n},$ the convex hull is defined as
$$\Conv(A) = \left\{\sum_{i=1}^m \alpha_i a_i : \forall i\ \alpha_i\in\mathbb{R}_{\ge0} \land \sum_i \alpha_i =1\right\},$$
where $a_i\in \mathbb{R}^n$ are the rows of $A$.
\end{definition}

\begin{theorem} (Stiemke's theorem e.g. \cite{dantzig1963linear})\\
\label{thm_stiemke}
Let $A\in\mathbb{R}^{m\times n}$ be a matrix, then the following two expressions are equivalent.
\begin{itemize}
\item $\nexists y: Ay\lneq0$
\item $\exists x>0:A^Tx=0$
\end{itemize}
Here $z\lneq0$ means that $0\ne z\le0$ .
\end{theorem}

\begin{theorem} (Singleton solutions of inequality systems)  \label{thm_singleton_solution_inequalities}\\
Let $A\in\mathbb{R}^{m\times n}$, $b\in\mathbb{R}^m$ and $x\in\mathbb{R}^n$.
Furthermore, let the inequality system $$Ax+b\le0,$$ written as $(A,b)$, have a solution $x_0$. \\
Then this solution is unique if and only if there exists an index set, $I$, for the rows s.t. $(A|_I, b|_I)$ is omnidirectional for $x_0$.
\end{theorem}

\begin{proof} (Theorem~\ref{thm_singleton_solution_inequalities}, Singleton solutions of inequality systems) \\
\underline{``$\Leftarrow$''}\\
Let $(A|_I,b|_I)$ be omnirectional for $x_0.$ Then it holds that $A|_Ix+b|_I=A|_I(x-x_0)\le0.$ Due to the omnidirectionality of $ A|_I, $ $x_0$ is the unique solution of the inequality system $ A|_Ix+b|_I\le0. $ The existence of a solution for the whole system $ Ax+b\le0 $ is guaranteed by assumption and therefore $ x_0 $ is the unique solution of $ Ax+b\le0. $ \\
\underline{``$\Rightarrow$''}\\
Here we will proove $$\text{``}\nexists I: (A|_I,b|_I)\text{ omnidirectional for some $p$ }\Rightarrow \text{ solution non-unique} \text{''}.$$

We will start by doing the following logical transformations:
\begin{align*}
& \phantom{\Leftrightarrow\ \ } \nexists I: (A|_I,b|_I)\text{ omnidirectional for some $p$ }\\
&\Leftrightarrow \nexists (I,p): (A|_I \text{ omnidirectional}\land b|_I=-A|_Ip) \\
&\Leftrightarrow \forall (I,p): \lnot (A|_I \text{ omnidirectional}\land b|_I=-A|_Ip) \\
&\Leftrightarrow \forall (I,p): (A|_I \text{ not omnidirectional}\lor b|_I\ne-A|_Ip).
\end{align*}

Now we define the vector $c_0 := Ax_0+b\le0$ and the set $I$ as the index set given via $c_0=0$.

This means that $A|_I$ is not omnidirectional, because otherwise $A|_Ix_0+b|_I=0$ due to the definition of $ I, $ which would lead to the contradiction that $(A|_I,b|_I)$ is omnidirectional for $x_0.$
But this means $\exists x'\ne0:A|_Ix'\le0$ as a result of Corollary \ref{corollary_omnidirectional}. Since $A|_{I^c}x_0+b|_{I^c}<0,$ we also have $\forall x \ \ \exists \epsilon>0:A|_{I^c}(x_0+\epsilon x)+b|_{I^c}<0.$ This holds in particular for $ x', $ so we define accordingly $ x^*:= \varepsilon x' \neq 0. $ Therefore, we have $ A|_{I^c}(x_0+x^*)+b|_{I^c}<0 $ as well as
\begin{equation*}
	A|_I(x_0 + x^*) + b_I = \underbrace{A|_I x_0 + b_I}_{=c_0 = 0} + \underbrace{\varepsilon A|_I x'}_{\leq 0} \leq 0.
\end{equation*}
Altogether it holds that $ A(x_0 + x^*) + b \leq 0 $ with $ x^*\neq 0, $ which means that $ x_0 $ is a non-unique solution for the inequality system $ Ax + b\leq 0. $
\end{proof}

\begin{proof}(Theorem \ref{theorem_pre-images}, Preimages of $ \relu $-layers) \\
We consider the $ \relu $-layer
\begin{equation*}
y=\relu(Ax+b),
\end{equation*}
given its output $y\in\mathbb{R}^m$ with $A\in\mathbb{R}^{m\times n}$, $b\in\mathbb{R}^m$ and $x\in\mathbb{R}^n.$ Clearly, this equation can also be written as the mixed linear system
\begin{align*}
A|_{y>0}x+b|_{y>0}&=y|_{y>0},\\
A|_{y=0}x+b|_{y=0}&\le0.
\end{align*}
This allows us to consider the two cases $$\sN(A|_{y>0})=\{0\} \text{ and }\sN(A|_{y>0})\ne\{0\}.$$
In the first case, we have a linear system which allows us to calculate $x$ uniquely, i.e. we can do retrieval. This leads us to the second case, the interesting one.
In this case we can only recover $x$ uniquely if and only if the system of inequalities ``pins down'' $P_{\sN(A|_{y>0})}x$, where $P_V$ is the orthogonal projection into the closed space $V$. Formally this requires 
\[A|_{y\le0}(P_{\sN(A|_{y>0})^\perp}x+P_{\sN(A|_{y>0})}x)+b|_{y\le0}\le0\] to have a unique solution for $x\in\mathbb{R}^n$ and $P_{\sN(A|_{y>0})^\perp}x$ fixed (given via the equality system).
By defining $\overline b := b|_{y\le0} + A|_{y\le0}(P_{\sN(A|_{y>0})^\perp}x)$ we have
\[A|_{y\le0}(P_{\sN(A|_{y>0})}x) + \overline b \le0.\]
If $O\in\mathbb{R}^{k\times n}$ now denotes an orthonormal basis of $\sN(A|_{y>0})$, where $k:=\dim \sN(A|_{y>0})$, we can write \begin{equation*}\overline{A}\overline{x}+\overline b \le0,\end{equation*} where $\overline A := AO^T$ and $\overline x := Ox$ is a general element in $\mathbb{R}^k$.
It now follows from Theorem\ref{thm_singleton_solution_inequalities} that the inequality system $(\overline A, \overline b)$ has a unique solution if and only if $(\overline A, \overline b)$ has a subset of rows that are omnidirectional for some point $p$.
\end{proof}


\begin{proof} (Theorem~\ref{thm_convex_hull}, Convex hull theorem) \\
Since $\sN(A)=\{0\}$ follows from both sides of the equivalence, the following sequence of equivalencies holds.
$0\in\interior{\Conv(A)}$ $\Leftrightarrow$ $\exists x>0:A^Tx=0$ $\xLeftrightarrow{\text{Theorem \ref{thm_stiemke}}}$ $\nexists y:Ay\lneq0$ $\xLeftrightarrow{\text{Corollary~\ref{corollary_omnidirectional} and $\sN(A)=\{0\}$}}$ $\nexists y\ne0:Ay\le0$
\end{proof}

\section{Proofs for section \ref{sec:stability}}

\begin{proof}(Proof of Lemma \ref{lem:globalBound}, Global upper bound for largest/smallest singular value)\\
The upper bound on the largest singular value is trivial, as $\relu$ is contractive or in other terms $\|D_I A x\|_2 \leq \|Ax\|_2$ for all $I$ and $x \in \mathbb{R}^n$. \\
To prove the upper bound for the smallest singular value, we assume 
\begin{align}
\label{eq:assumBound}
\sigma_M := \min \{\sigma_l:  \sigma_l >0 \} > \tilde{\sigma}_k
\end{align}
and aim to produce a contradiction. Consider all singular vectors $\tilde{v}_{k^*}$ with $k^* \geq k$ from matrix $A$. It holds for all $\tilde{v}_{k^*}$
\begin{align}
\label{eq:projEst}
\tilde{\sigma}_k \geq \tilde{\sigma}_{k^*} = \|A\tilde{v}_{k^*}\|_2 \geq \|D_I A \tilde{v}_{k^*}\|_2,
\end{align}
as $D_I$ is a projection matrix and thus only contracting. As 
\begin{align*}
\sigma_M = \min_{\substack{\|x\|_2=1 \\ x \in \mathcal{N}(D_I A)^\bot}} \|D_I A x\|_2,
\end{align*}
all $\tilde{v}_{k^*} \not \in \mathcal{N}(D_I A)^\bot$. Otherwise, a $\tilde{v}_{k^*}$ would be a minimizer by estimation \eqref{eq:projEst}, which would violate the assumption \eqref{eq:assumBound}.\\
\\
Due to $\mathcal{N}(D_I A)^\bot \oplus \mathcal{N}(D_I A) = \mathbb{R}^n$, it holds $\tilde{v}_{k^*} \in \mathcal{N}(D_I A)$. As $\tilde{v}_{k^*}$ are orthogonal, $\text{dim}(span(v_{k^*})) = |I| + 1$ (note: $k^*=k, ..., N$ and $k= N-|I|$, thus there are $|I|+1$ singular vectors $v_{k^*}$ in total). Furthermore, $\tilde{v}_{k^*}$ were not in $\mathcal{N}(A)$ by definition (corresponding singular values were strictly positive).\\
Hence, the nullspace of $D_I$ must have $\text{dim}(\mathcal{N}(D_I)) \geq |I| + 1$. But $D_I$ is the identity matrix except $|I|$ zeros on the diagonal, thus $\text{dim}(\mathcal{N}(D_I)) = |I|$, which yields a contradiction.
\end{proof}

\begin{proof}(Proof of Lemma \ref{lem:corrRows}, Weakly correlated rows)\\
Consider $v = \frac{a_k}{\|a_k\|_2}$. Then,
\begin{align*}
(D_I A v)_k = 0,
\end{align*}
since $k \in I$ ($k$-th row of $D_I$ is zero). Furthermore, for all $j \neq k$ it holds by condition \eqref{eq:corrCond}
\begin{align*}
(D_I A v)_j = \frac{\langle a_k, a_j\rangle}{\|a_k\|_2} \leq \frac{|\langle a_k, a_j\rangle|}{\|a_k\|_2} \leq \frac{c}{\sqrt{M}}.
\end{align*}
Hence,
\begin{align*}
\|D_I A v\|_2 = \sqrt{\sum_{j \not \in I} \left(\frac{\langle a_k, a_j\rangle}{\|a_k\|_2}\right)^2} \leq \sqrt{M \left(\frac{c}{\sqrt{M}}\right)^2} = c. 
\end{align*}
As $a_k \in \mathcal{N}(D_I A)^\bot$, $v \in \mathcal{N}(D_I A)^\bot$ as well. Thus,
\begin{align*}
\sigma_K = \min_{\substack{\|x\|_2 \\ x \in \mathcal{N}(D_I A)^\bot}} \|D_I A\|_2 \leq \|D_I A v\|_2 \leq c.
\end{align*}
\end{proof}

\newpage
\section{Appendix for Section~\ref{sec:retrievalnumerics}}
\label{app:algoRetrieval}
In this section, we formulate the algorithm to determine whether the preimage of $y$ given by $$y=\relu(Ax+b)$$ is finite.\\
This requires to check whether $\overline A$ (see Theorem~\ref{theorem_pre-images}) is omnidirectional, which is equivalent to 
\begin{align*}
0\in\interior{\Conv(\overline A)},
\end{align*}
see Theorem \ref{thm_convex_hull}. Since it is reasonable to assume that $0$ will not lie on the boundary of the convex hull, we can formulate this as a \textit{linear programming} problem. The side-conditions incorporate the definition of convex hulls (Definition \ref{def_convex_hull}, Appendix \ref{app:uniqueness}). The objective function is chosen arbitrary, as we are only interested in a solution.

\begin{algorithm}[h!]
   \caption{Finite preimage}
   \label{alg:preimage}
\begin{algorithmic}
   \STATE {\bfseries Input:} $A\in\mathbb{R}^{m\times n}$, $b\in\mathbb{R}^m$, $y\in\mathbb{R}^m$
   \IF{$\rank(A|_{y>0})=n$}
   \STATE \textbf{return} True \COMMENT{Preimage is a singleton}
   \ENDIF
   \STATE $O \gets$ orthonormal basis of $\sN(A|_{y>0})$, ($\in\mathbb{R}^{k\times n}$)
   \STATE $\overline A \gets A|_{y=0}O^T$, ($\in\mathbb{R}^{\tilde k\times k}$)
   \IF{$\tilde k\le k$}
   \STATE \textbf{return} False \COMMENT{see Corollary~\ref{corollary_omni_set_lines}}
   \ENDIF
   \STATE $c\gets(1;\dots;1)$ \COMMENT{arbitrary objective}
   \STATE \textbf{return} Does a solution for the linear program 
   		$\begin{cases}
			\mbox{max}\ c^Tx \\
			\text{subject to} \\
			\hspace{1cm} \overline A^Tx=0 \\
			\hspace{1cm} (1;\dots;1)^T x = 1 \\
			\hspace{1cm} x\in[0,1]^{\tilde k}
		\end{cases}$
		exists?
\end{algorithmic}
\end{algorithm}

\section{Architectures for Numerical Studies}
\label{app:architectures}
Training details for MLP on MNIST:
\begin{itemize}
\item Training using Adam optimizer \citep{adam}
\item Epochs: 25
\item Batch size: 1000
\end{itemize}
Training details for WideCIFAR and ThinCIFAR:
\begin{itemize}
\item Training setup from Keras \citep{keras} examples: \texttt{cifar10\char`_cnn}
\item No data augmentation
\item RMSprop optimizer
\item Epochs: 100
\item Batch size: 32
\end{itemize}

\begin{table}[!h]
\centering
\caption{Architecture of MLP trained on MNIST \label{Tab:mlpMNIST_opening_images} \vspace{0.1cm}} {
\begin{tabular}{@{}clcccc@{}}\toprule 
Index & Type  & kernel size & stride & \# feature maps & \# output units\\\midrule
0 & Input layer  & - & - & 3 \\
1 &Dense layer & - & - & - & 100 \\ 
2 &Dense layer & - & - & - & 100 \\ 
3 &Dense layer & - & - & - & 100 \\ 
4 &Dense layer & - & - & - & 100 \\ 
5 &Dense layer & - & - & - & 100 \\ 
6 &Dense layer & - & - & - & 100 \\ 
7 &Dense layer & - & - & - & 100 \\ 
8 &Dense layer & - & - & - & 100 \\ 
9 &Dense layer & - & - & - & 100 \\ 
10 &Dense layer & - & - & - & 100 \\ 
11 &Dense layer (softmax) & - & - & - & 10\\
\midrule
\end{tabular}}{}
\end{table}

\begin{table}[!h]
\centering
\caption{Architecture of MLP trained on MNIST \label{Tab:mlpMNIST} \vspace{0.1cm}} {
\begin{tabular}{@{}clcccc@{}}\toprule 
Index & Type  & kernel size & stride & \# feature maps & \# output units\\\midrule
0 & Input layer  & - & - & 3 \\
1 &Dense layer & - & - & - & 3500 \\ 
2 &Dense layer & - & - & - & 784 \\ 
3 &Dense layer (softmax) & - & - & - & 10\\
\midrule
\end{tabular}}{}
\end{table}

\begin{table}[!h]
\centering
\caption{Architecture of WideCIFAR \label{Tab:wideCifar} \vspace{0.1cm}} {
\begin{tabular}{@{}clcccc@{}}\toprule 
Index & Type  & kernel size & stride & \# feature maps & \# output units\\\midrule
0 & Input layer  & - & - & 3 \\
1 & Convolutional layer  & (3,3) & (1,1) & 32 &-\\
2 & Convolutional layer  & (3,3) & (2,2) & 64 &-\\
3 & Convolutional layer  & (3,3) & (1,1) & 64 &-\\
4 & Convolutional layer  & (3,3) & (1,1) & 32 & -\\
5 & Convolutional layer  & (3,3) & (1,1) & 32 & -\\
6 & Convolutional layer  & (3,3) & (2,2) & 64 &-\\
7 & Dense layer & - & - & - & 512 \\ 
8 & Dense layer (softmax) & - & - & - & 10\\
\midrule
\end{tabular}}{}
\end{table}

\begin{table}[!h]
\centering
\caption{Architecture of ThinCIFAR \label{Tab:thinCifar} \vspace{0.1cm}} {
\begin{tabular}{@{}clcccc@{}}\toprule 
Index & Type  & kernel size & stride & \# feature maps & \# output units\\\midrule
0 & Input layer  & - & - & 3 \\
1 & Convolutional layer  & (3,3) & (1,1) & 32 & -\\
2 & Convolutional layer  & (3,3) & (2,2) & 32 & -\\
3 & Convolutional layer  & (3,3) & (1,1) & 16 & -\\
4 & Convolutional layer  & (3,3) & (1,1) & 16 &- \\
5 & Convolutional layer  & (3,3) & (1,1) & 16 & -\\
6 & Convolutional layer  & (3,3) & (2,2) & 32 & -\\
7 &Dense layer & - & - & - & 512 \\ 
8 &Dense layer (softmax) & - & - & - & 10\\
\midrule
\end{tabular}}{}
\end{table}

\newpage
\section{Numerical Analysis of Lemma \ref{lem:corrRows}}
\label{app:corrAnalysis}
\begin{figure}[!h]
    \begin{center}
    \includegraphics[width=0.5\textwidth]{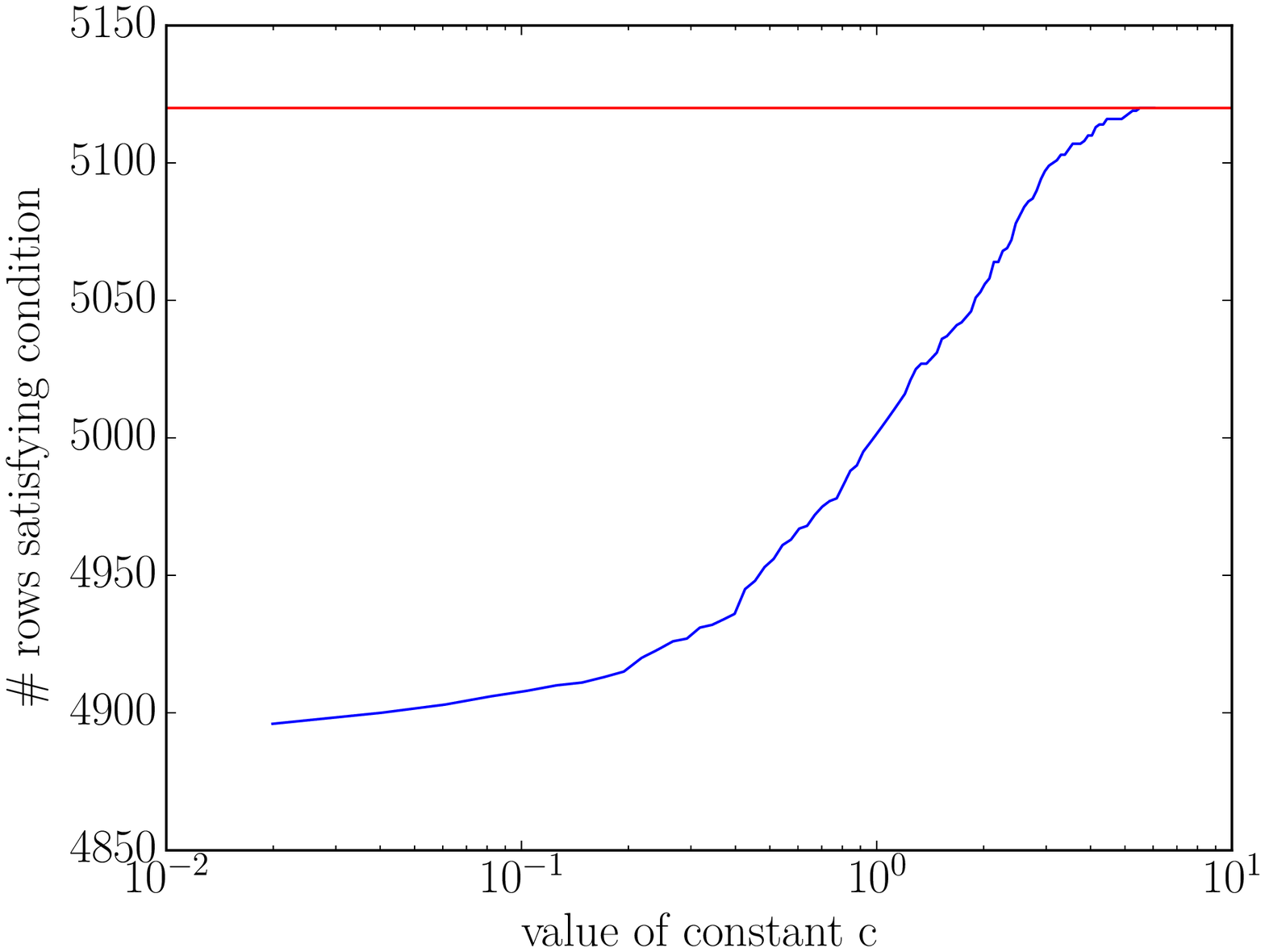} 
    \end{center}
\caption{Curve showing how many rows $a_i$ satisfy condition \eqref{eq:corrCond} from Lemma \ref{lem:corrRows} depending on values of constant $c$. The red line shows the total number of remaining rows after removal by $\relu$, $M=5120$. Even for small constants $c$ most $a_i$ fulfill condition \eqref{eq:corrCond}, yet not all, which is required by the lemma to give an upper bound on the smallest singular value. The example is from layer 4 of WideCIFAR, for only one sample from the test set.}
\label{fig:corrBound}
\end{figure}
\begin{figure}[!h]
\label{fig:thinCifar}
\begin{minipage}[c]{0.5\textwidth}
    \centering
\includegraphics[width=\textwidth]{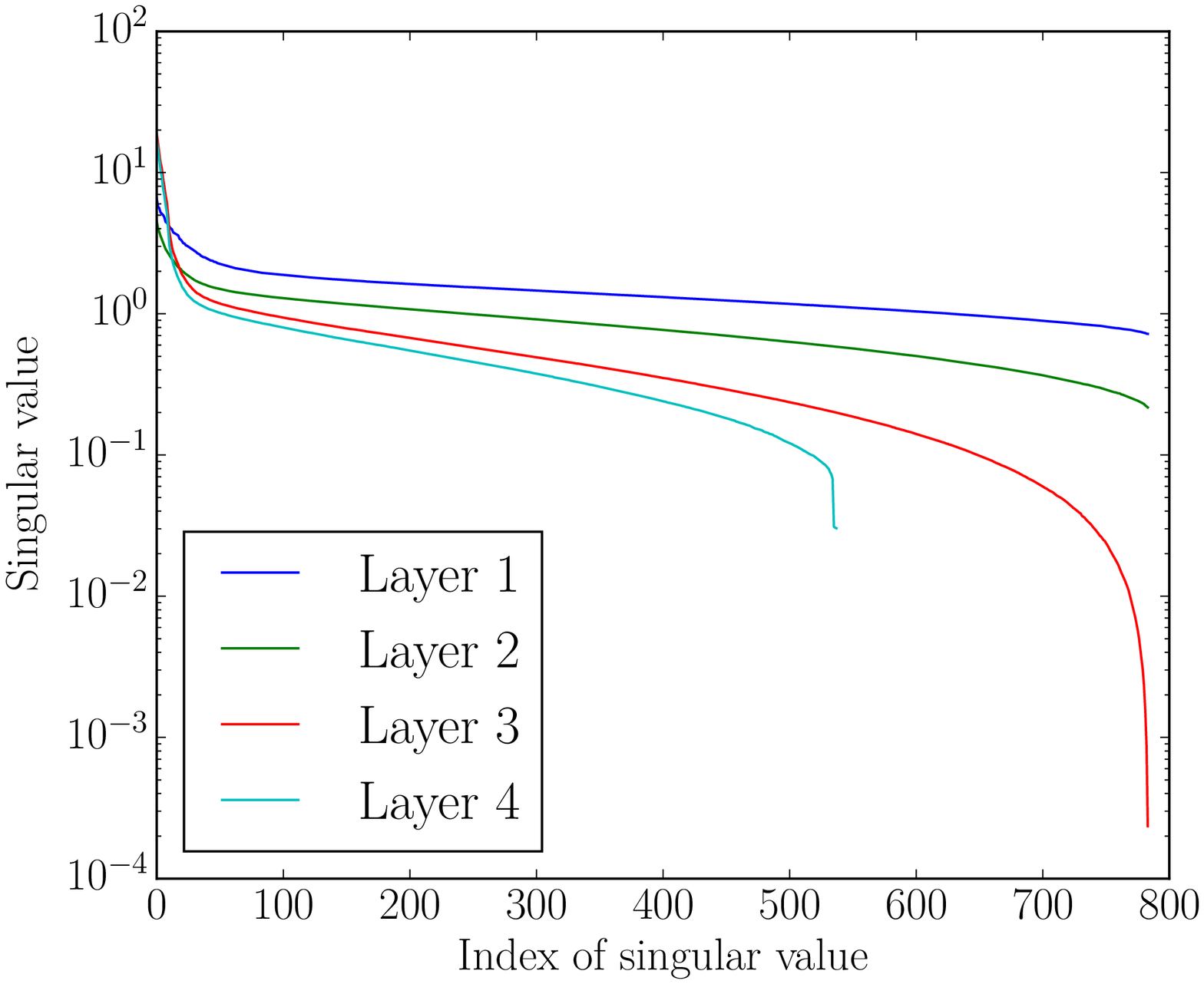} 
\end{minipage}
\begin{minipage}[c]{0.5\textwidth}
    \centering
\includegraphics[width=\textwidth]{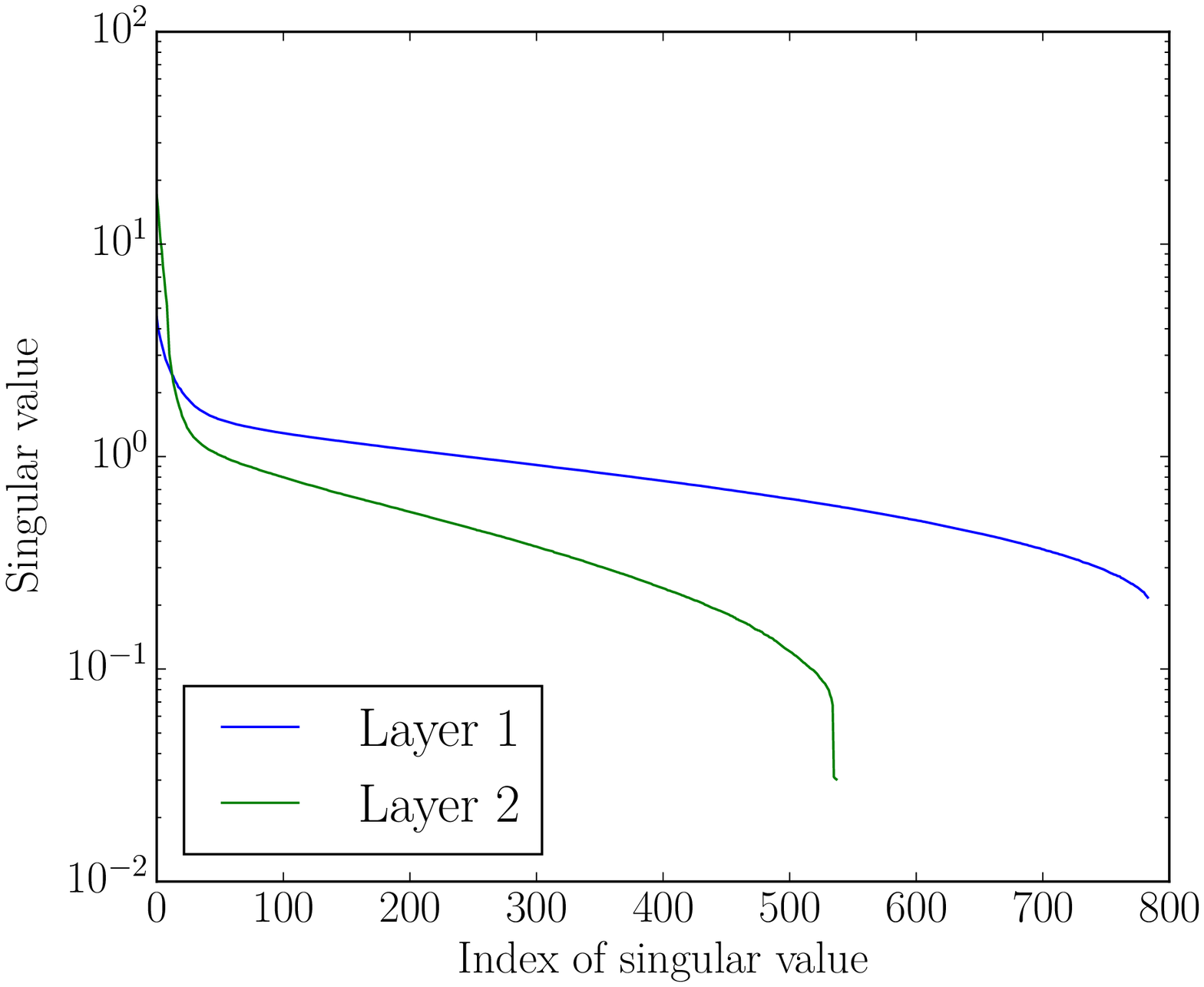} 
\end{minipage}
\caption{Left:  Effect of $\relu$ on the singular values for the MLP on MNIST. The curves show the effect in layer 1(layer 1 and 2 in legend, because $\relu$ is counted as an extra activation layer) and layer 2 (layer 3 and 4). Right: Decay of singular values over the layers of MLP on MNIST. Here, each layer includes the fully-connected layer and $\relu$-activation. Reported number are taken from median over 50 samples.}
\label{fig:MLPsing}
\end{figure}

In order to better understand the bound on the smallest singular value after $\relu$, given by Lemma \ref{lem:corrRows}, we numerically proceed as follows:
\begin{enumerate}
\item We choose $c \in [a, b]$, where $a,b$ are suitable interval endpoints.
\item Given $c$, we compute for every $a_k$ with $k \in I$ the value of $c \frac{\|a_k\|_2}{\sqrt{M}}$ ($M$ is the number of remaining rows, in the example $M=5120$).
\item For every $a_k$ we count the number of $a_i$ satisfying
\begin{align*}
|\langle a_i, a_k\rangle| \leq c \frac{\|a_k\|_2}{\sqrt{M}}.
\end{align*}
\item We take the $a_k$ with the maximal number of $a_i$ satisfying the condition. (Note, that this ignores the requirement $a_k \in \sN(D_I A)^\bot$.)
\item If we have an $a_k$, where all $a_i$ satisfy the condition, the corresponding constant $c$ gives the upper bound on the smallest singular value after $\relu$.
\end{enumerate}
Figure \ref{fig:corrBound} shows the number of $a_i$ satisfying the correlation condition given different choices of $c$. The red line is reached for $c\approx 6$. However, even the largest singular value after $\relu$ is smaller than 2.5 (shown in Figure \ref{fig:effectRelu}). Thus, the bound given by Lemma \ref{lem:corrRows} is far off. This can be explained by the fact, that this situation is quite redundant ($M=5120$) and there are rows $a_i$ still correlated to the removed rows $a_k$. \\
However, in the further Experiments on ThinCIFAR, we observe (see Figure \ref{fig:thinCifar}) a stronger effect of $\relu$ in layer 2, which can be explained by having a less redundant scenario with fewer remaining rows.

\newpage
\newpage
\section{Further Experiments}
\label{app:furtherExp}
\begin{figure}[!h]
\begin{minipage}[c]{0.5\textwidth}
    \centering
\includegraphics[width=\textwidth]{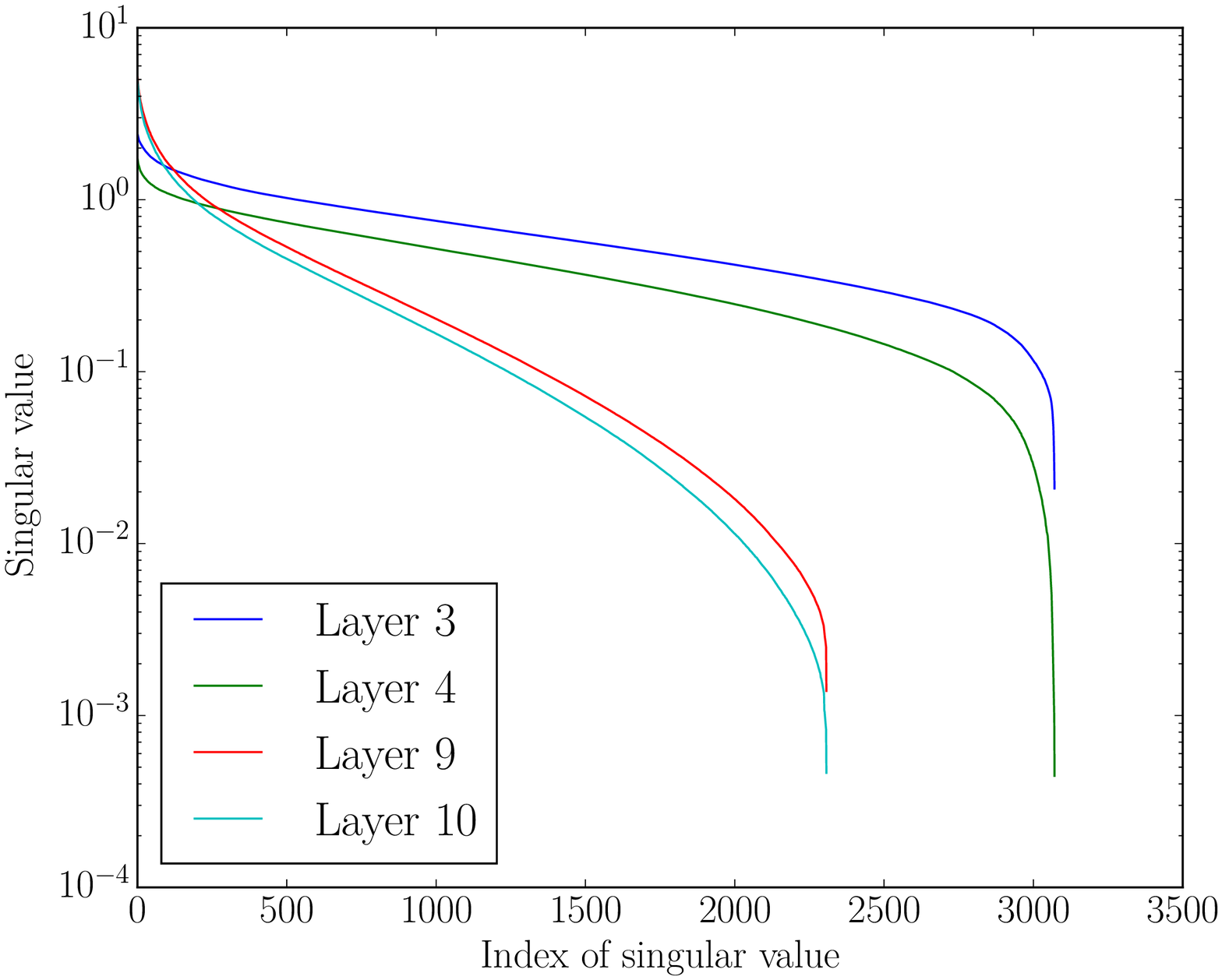} 
\end{minipage}
\begin{minipage}[c]{0.5\textwidth}
    \centering
\includegraphics[width=\textwidth]{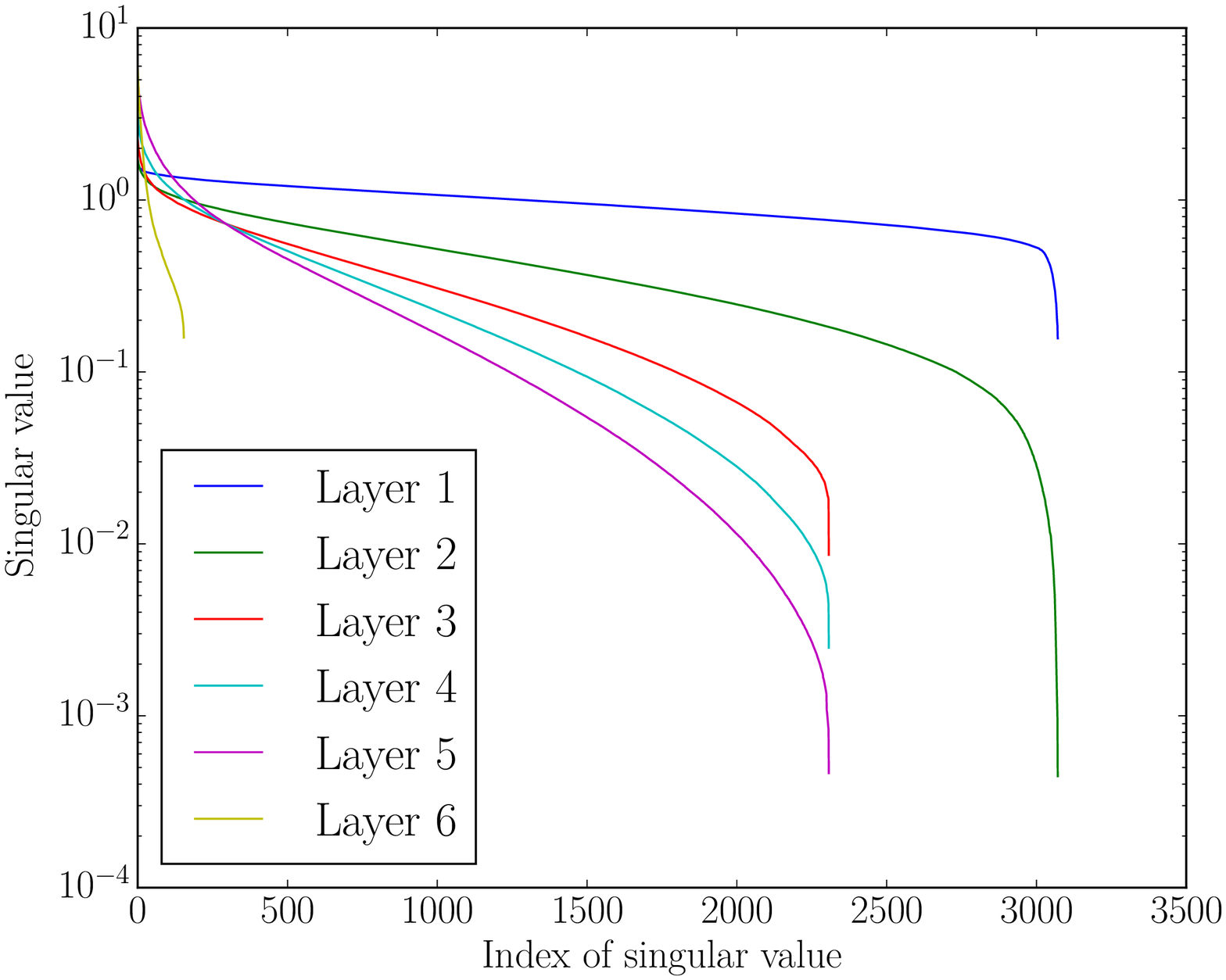} 
\end{minipage}
\caption{Left: Effect of $\relu$ on the singular values for ThinCifar. The curves show the effect in layer 2 (layer 3 and 4 in legend, because $\relu$ is counted as an extra activation layer) and layer 5 (layer 9 and 10). Right: Decay of singular values over the layers ThinCifar. Here, each layer includes the convolution and $\relu$-activation. Reported number are taken from median over 50 samples. Best viewed in color.}
\end{figure}

\section{Invariance Experiment using an MLP on MNIST}
\label{app:invarianceMNIST}
This section briefly describes how the results in Figure \ref{opener_image} from the introduction were obtained (copied in Figure \ref{opener_images_copy} for readability). After training the network from \ref{Tab:mlpMNIST_opening_images} (in Appendix \ref{app:architectures}), we searched the MNIST test set for input images with yielded the fewest positive activations in the first layer, in the figure the digits "3" and "4". After selecting the example input $x^*$, we selected another input $c$ belonging to a different class (e.g. a "6" and "4" in the first example).\\
Afterwards, we solved following linear programming problem to find a perturbed $x$:
\begin{align*}
\begin{cases}
			\mbox{max}\ \langle c, x\rangle \\
			\text{subject to} \\
			\hspace{1cm}  A|_{y^* >0} x + b|_{y^*>0}=y^*|_{y^*>0} \\
			\hspace{1cm}  A|_{y^* <0} x + b|_{y^*<0}\leq 0 \\
			\hspace{1cm} x\in[0,1]^{\tilde k}
		\end{cases},
\end{align*}
where the features of the first layer are computed via
\begin{align*}
y^* = \relu(Ax^* + b).
\end{align*}
Hence, we searched within the preimage of the features $y^*$ of the first layer for examples $x$ which resemble images $c$ from another class. By doing this we observe, that the preimages of the MLP may have large volume. In these cases, the network is invariant to some semantics changes which shows how the study of preimages can reveal previously unknown properties.\\

\begin{figure}
\includegraphics[width=0.485\textwidth]{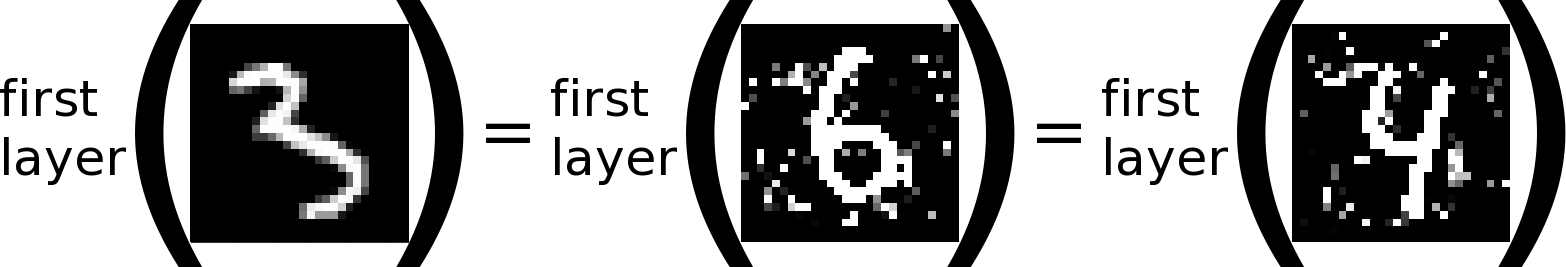}
\hfill
\includegraphics[width=0.485\textwidth]{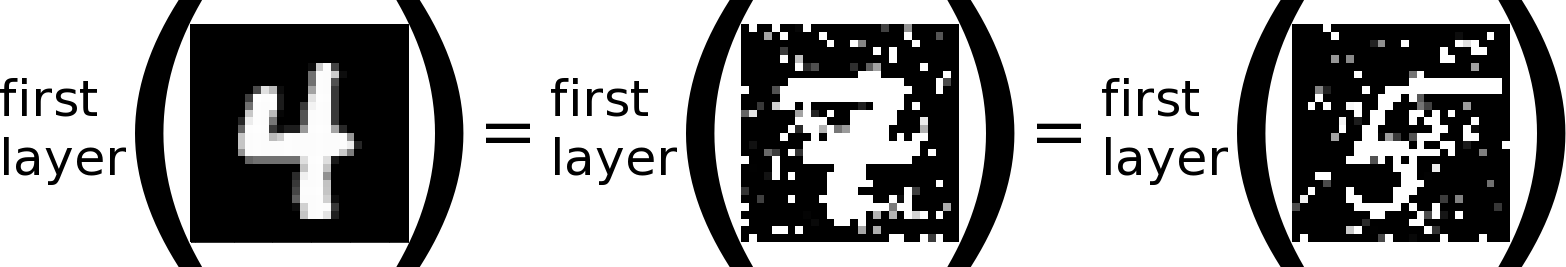}
\caption{Invariances of the first layer (100 $\relu$ neurons) of a vanilla MLP. (Exact architecture in Appendix~\ref{app:architectures} Table \ref{Tab:mlpMNIST_opening_images}.) }
\label{opener_images_copy}
\end{figure}

\end{document}